\pdfoutput=1

\documentclass[11pt]{article}
\usepackage{tabularx}
\usepackage{booktabs} 
\usepackage[final]{acl}
\usepackage{graphicx} 
\usepackage{caption} 
\usepackage{booktabs} 
\usepackage{listings}
\usepackage[justification=centering]{caption}

\usepackage{times}
\usepackage{latexsym}
\usepackage{booktabs}
\usepackage{adjustbox}
\usepackage{graphicx}
\usepackage{float}
\usepackage{url}
\usepackage{placeins}
\usepackage[T1]{fontenc}
\usepackage[justification=justified]{caption}

\usepackage[utf8]{inputenc}

\usepackage{microtype}
\usepackage{multirow}
\usepackage{inconsolata}

\usepackage{graphicx}
\usepackage{amsmath}
\usepackage{caption}
\usepackage{amssymb}
\usepackage{pifont}
\title{
    \raisebox{-0.2cm}{\includegraphics[width=0.05\textwidth]{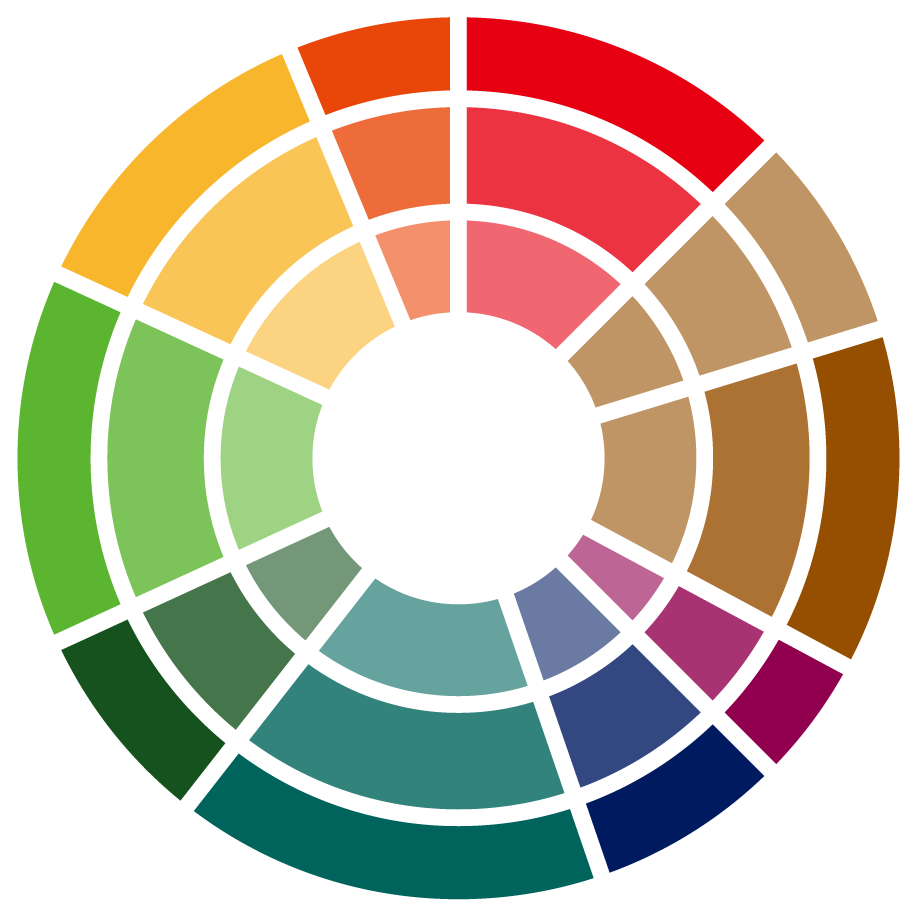}}  
    \hspace{-0.1cm} 
    Value-Spectrum: Quantifying Preferences of Vision-Language Models via Value Decomposition in Social Media Contexts
}

\author{
\textbf{Jingxuan Li\textsuperscript{1,*}} \quad
\textbf{Yuning Yang\textsuperscript{1,*}} \quad
\textbf{Shengqi Yang\textsuperscript{2}} \quad
\textbf{Linfan Zhang\textsuperscript{1}} \quad
\textbf{Ying Nian Wu\textsuperscript{1}} \\
\textsuperscript{1}University of California, Los Angeles \quad
\textsuperscript{2}Los Alamos National Laboratory \\
\texttt{\{madili, yuningyang, linfanz\}@g.ucla.edu} \quad
\texttt{shengqi@lanl.gov} \quad
\texttt{ywu@stat.ucla.edu} \\
}

\begin{document}
\maketitle
\renewcommand{\thefootnote}{\fnsymbol{footnote}} 
\footnotetext[1]{Equal contribution, alphabetical by first name.}


\begin{abstract}
The recent progress in Vision-Language Models (VLMs) has broadened the scope of multimodal applications. However, evaluations often remain limited to functional tasks, neglecting abstract dimensions such as personality traits and human values. To address this gap, we introduce Value-Spectrum, a novel Visual Question Answering (VQA) benchmark aimed at assessing VLMs based on Schwartz's value dimensions that capture core human values guiding people’s preferences and actions. We design a VLM agent pipeline to simulate video browsing and construct a vector database comprising over 50,000 short videos from TikTok, YouTube Shorts, and Instagram Reels. These videos span multiple months and cover diverse topics, including family, health, hobbies, society, technology, etc. Benchmarking on Value-Spectrum highlights notable variations in how VLMs handle value-oriented content. Beyond identifying VLMs' intrinsic preferences, we also explore the ability of VLM agents to adopt specific personas when explicitly prompted, revealing insights into the adaptability of the model in role-playing scenarios. These findings highlight the potential of Value-Spectrum as a comprehensive evaluation set for tracking VLM preferences in value-based tasks and abilities to simulate diverse personas. The complete code and data are available at \url{https://github.com/Jeremyyny/Value-Spectrum}.

\end{abstract}

\section{Introduction}
\begin{figure}[t]
  \includegraphics[width=\columnwidth]{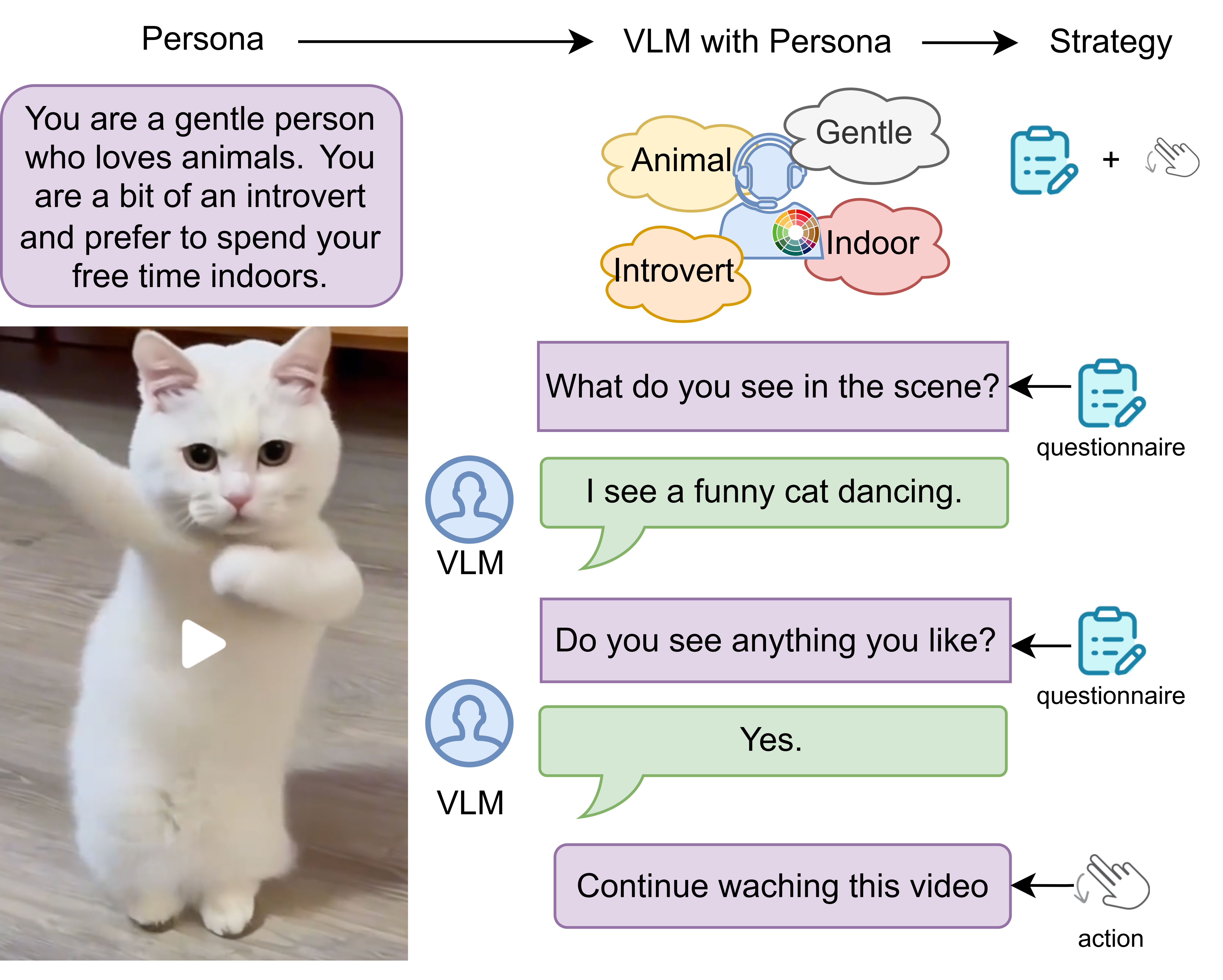}
   \caption{\textbf{Exploring value-driven role-playing in VLMs.} This study investigates how VLMs adopt assigned personas to align value traits and preferences within social media contexts.}
  \label{fig:thumb}
\end{figure}

\begin{figure*}
    \centering
    \includegraphics[width=1\textwidth]{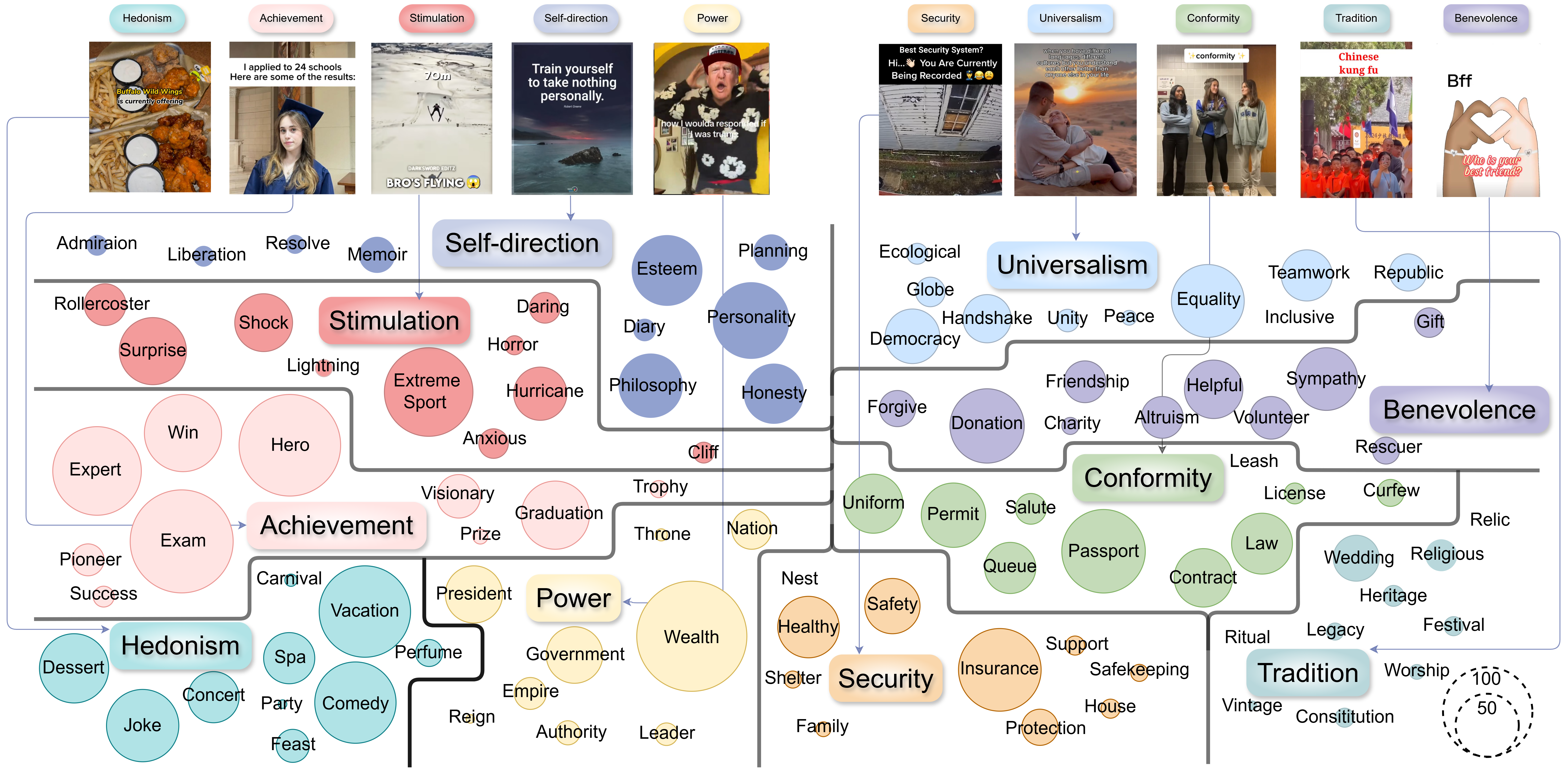}
    \caption{\textbf{Overview of short video contents distribution of \textit{Value-Spectrum} dataset.} We collect an abundance of short video screenshots relevant to 10 Schwartz values. The area of circles centered at each keyword represents the amount of relevant videos in the dataset.}
    \label{fig:valuenet}
\end{figure*}

Vision-Language Models, built upon Large Language Models (LLMs) with pre-trained vision encoders through cross-modal alignment training, have shown impressive perceptual and cognitive capabilities in tasks like VQA and image captioning~\cite{Zhou2019UnifiedVP, Radford2021LearningTV, Zhang2023VisionLanguageMF, Lu2024WildVisionEV}. Recent research has identified that LLMs exhibit distinct preferences~\cite{Li2024DissectingHA}, personalities~\cite{serapio2023personality}, and values~\cite{Ren2024ValueBenchTC}. In addition, some studies have explored the potential of LLMs as role-playing agents to simulate various personas~\cite{Wang2023InCharacterEP, Chen2024FromPT}.
Questions thus arise about whether VLMs, as visual extensions of LLMs, also exhibit inherent preferences and whether they can be induced to role-play specific personas.

To address these concerns, our study explores two key questions: (1) Do VLMs exhibit preference traits? (2) Can VLMs adapt their traits to role-play specific human-designed personas, aligning their behaviors and preferences to match predefined roles? To answer these questions, we propose a framework that systematically evaluates VLM preference traits through analyzing their values, i.e., the guiding principles that influence (human) attitudes, beliefs, and traits~\cite{schwartz2012overview}. By evaluating how VLMs prioritize these values, we can gain insights into their preference traits and alignment with human-designed personas.

In this paper, we introduce \textit{\textbf{Value-Spectrum}}, a benchmark designed to evaluate preference traits in VLMs through visual content from social media. Our framework utilizes VLM agents embedded within social media platforms to collect a dataset of $50,191$ unique short video screenshots spanning a wide range of topics, including lifestyle, technology, health, and more. To enable scalable evaluation, we construct a vector database using the CLIP model ~\cite{Radford2021LearningTV}, facilitating keyword-driven retrieval of images aligned with specific value dimensions. These images representing each value dimension are then presented to VLMs alongside designed questionnaires to assess their preferences.

\begin{figure*}
    \centering
    \includegraphics[width=1\textwidth]{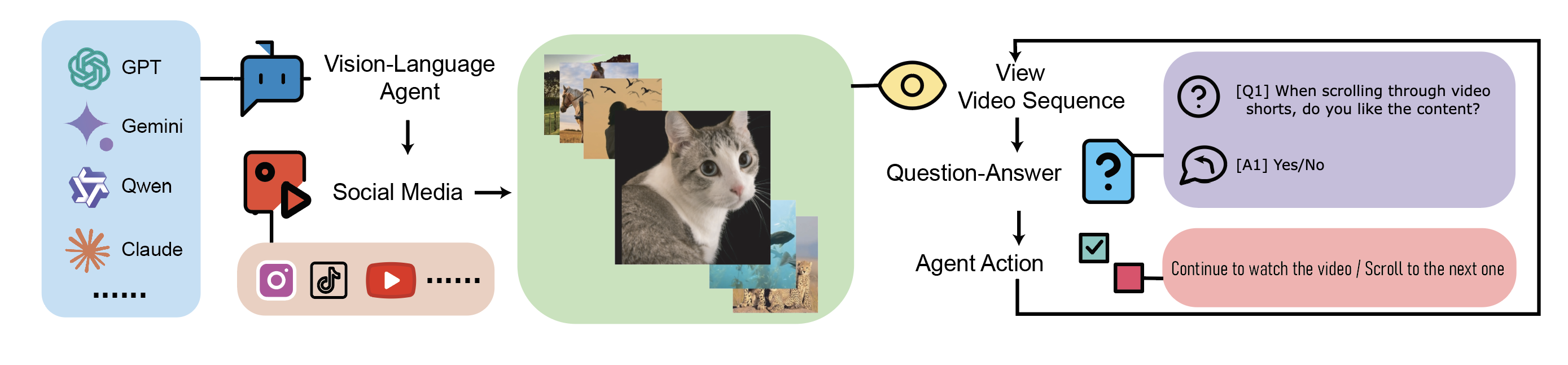}
    \caption{\textbf{Illustration of our VLM agent interaction with social media videos and screenshot collection pipeline} Our pipeline uses a VLM agent to capture screenshots from social media videos (e.g., Instagram, TikTok, YouTube). These screenshots are fed to a VLM (e.g., GPT, Gemini, Qwen, Claude), which then indicates its preference (Yes/No) for the content (e.g., in response to "Do you like this content?"). The agent subsequently acts (e.g., continues viewing or scrolls to the next video) based on this VLM feedback, enabling automated assessment of VLM responses to visual social media content.}
    \label{fig:pipeline}
\end{figure*}

Our findings reveal models' shared tendency to exhibit a strong inclination towards \textit{Universalism} and \textit{Benevolence}, but preferences still vary across models. CogVLM2 and Gemini 2.0 Flash demonstrate relatively balanced and high preferences across all value dimensions, while other models like GPT-4o and Claude 3.5 Sonnet show distinct preferences, favoring values like \textit{Universalism} and dislikes \textit{Stimulation}. In contrast, Blip-2 shows low engagement across most value dimensions, with the highest standard deviation, indicating a random preference pattern with inconclusive responses for reasons of likes and dislikes.

In addition to the static preferences of VLMs, we evaluate their abilities to adapt inherent preferences in role-playing specific personas. We propose two strategies, \textit{Simple} and \textit{Inductive Scoring Questionnaire(ISQ)}, to assess the effectiveness of different prompt techniques in inducing VLMs with injected personas. By evaluating these strategies across multiple platforms, our experiments show that TikTok serves as an optimal testing environment for inducing VLM personalities, with models demonstrating the strongest alignment under the ISQ strategy. Notably, Claude 3.5 Sonnet achieved the highest alignment with the ISQ strategy, whereas Blip-2 showed no preference alignment under either strategy, underscoring fundamental differences in model adaptability.

This work makes the following contributions:
\begin{itemize}
  \item We propose \textit{\textbf{Value-Spectrum}}, a benchmark for quantifying VLM value preferences, using social media-based contents to reveal human-like traits across different VLMs.
    \item We present a dataset of over $50$k short video screenshots spanning diverse topics, social media platforms, and release dates, designed for our benchmark.
  
    \item We embed specific role-play personas into VLMs using two strategies(Simple and ISQ) to adjust their value traits, achieving improved personality alignment in real-world interactions.
\end{itemize}

\section{Related work}

\subsection{Vision-Language Agents}

VLMs take inputs as images and textual descriptions, and they learn to discover the knowledge from the two modalities. The recent development of large VLMs is rapidly advancing the field of AI. These models have the potential to revolutionize various industries and tasks, showcasing their power in plot and table identifying~\citep{liu2022deplot},VQA~\citep{hu2024bliva}, image captioning~\citep{bianco2023improving}, and e.t.c. Following \citet{niu2024screenagent}, the environment for VLM agents to interact with social media can be constructed. We design an automated control pipeline that guides the
agent to continuously interact with social networks. 

\subsection{Computational Social Science}

      The intersection of social media and computational social science has emerged as a dynamic field of research~\citep{chen2023using}. Dialogues and social interactions, with their vast user base and intricate networks of connections, offer a large database for studying human behaviors~\citep{christakis2013social}, social relationships~\citep{qiu2021socaog}, and social networks~\citep{zhang2023adjusted}. Researchers in computational social science apply advanced computational techniques, such as machine learning, natural language processing (NLP), and network analysis, to analyze massive datasets extracted from social media platforms. These analyses provide insights into various phenomena, including information diffusion~\citep{jiang2014evolutionary}, opinion formation~\citep{xiong2014opinion}, and collective behavior~\citep{pinheiro2016linking}.

\subsection{Sentiment, Personality, and Value}

The community has employed machine learning-based models to study human sentiment~\citep{malviya2020machine}, personality~\citep{stachl2020personality}, and values~\citep{qiu2022valuenet, Ye2024MeasuringHA}. Building on this, research has explored AI personalities, including those of LLMs~\citep{serapio2023personality, Li2023TheSO, Rttger2024PoliticalCO}, and their alignment with human values~\citep{ren2024valuebench}. Inspired by recent efforts to reveal AI agent traits through indirect methods like physiological exams~\citep{jiang2024evaluating}, questionnaires~\citep{huang2023emotionally}, and cultural views~\citep{kovavc2023large}, we adopt a novel perspective: examining machine behavior and inferred personality traits, including underlying values, to evaluate the performance of VLMs on mainstream social media platforms. To ground this examination, we utilize Schwartz’s value theory~\citep{schwartz1992universals}, a comprehensive and culturally validated classification of human motivations~\citep{davidov2008measuring} that predicts attitudes and social behavior~\citep{bardi2003values}. While prior computational work like ValueNet~\citep{qiu2022valuenet} and ValueBench~\citep{ren2024valuebench} has assessed values in text-based NLP systems, and recent studies have begun exploring cultural value sensitivity in multimodal models~\citep{Yadav2025BeyondWE}, our work specifically extends this value-based evaluation to VLMs within visual contexts on social media.

\begin{figure*}
    \centering
    \includegraphics[width=0.98\textwidth]{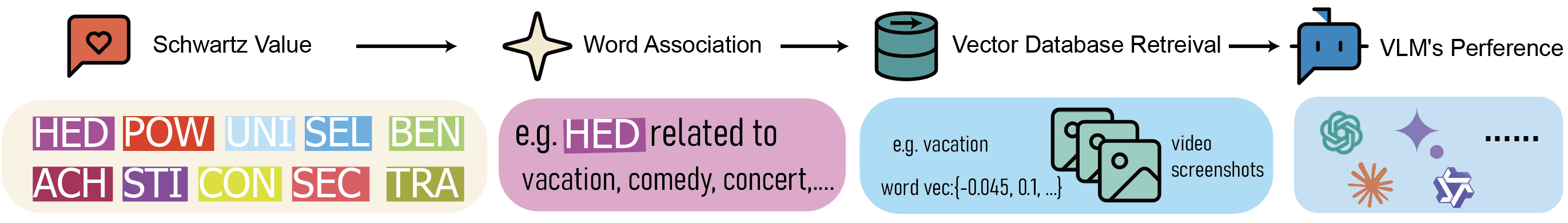}
    \caption{\textbf{Schwartz value-based image retrieval pipeline.} Our pipeline retrieves video screenshots based on Schwartz values by associating each value with relevant keywords, such as linking \textit{Hedonism} to topics like vacation. These keywords are converted into vector queries, retrieving matching video content using the shortest distance within our database.}
    \label{fig:retrieval}
\end{figure*}

\section{Data Collection} \label{sec:dataset}
Inspired by ScreenAgent~\cite{niu2024screenagent}, our work leverages a VLM-driven graphical user interface (GUI) agent to autonomously navigate popular social media platforms. This agent conducts random walks through social media platforms where it observes and captures screenshots. The data collected is stored in a vector database~\cite{han2023comprehensive}, creating a structured repository optimized for value decomposition and efficient retrieval. We aim to analyze VLMs' behaviors across diverse social contexts and reveal their preferences. The automated data collection process (see Figure~\ref{fig:pipeline}) efficiently fetches a large volume of diverse content, granting the analysis with scope and depth that traditional manual collection methods could not achieve.

The resulting dataset comprises $50,191$ videos sourced from Instagram ($32\%$), YouTube ($29\%$), and TikTok ($39\%$). Each entry includes the video link, a screenshot, and meta-information such as platform name and post date, capturing a comprehensive snapshot of content ranged between July 31, 2024, and October 31, 2024. By collecting data proportionally across these platforms, we enable balanced analysis and facilitate unbiased value decomposition across social media content. This innovative dataset empowers researchers to explore the behavior of VLMs in a systematic and organized way, fostering deeper insights into model interpretation and the dynamics of social media.

To examine the distribution of video themes in this dataset, we take a screenshot of each video at the beginning as a representation of the video content. In Figure~\ref{fig:valuenet}, we present the distribution of videos that are relevant to ten Schwartz values. Specifically, for each Schwartz value, we curate 10 representative keywords. The area of the transparent circle is proportional to the number of videos that lie within a distance of $1.5$ to the corresponding normalized keyword vector. We find that videos relevant to these Schwartz Value Dimensions \textit{Achievement}, \textit{Hedonism}, and \textit{Power} appear most frequently, while videos about \textit{Tradition} are relatively rare. We then vectorize the image and define its relevance to a specific keyword using cosine distance.\par

\section{Evaluating VLM's Preferences}

Extending the idea of analyzing LLMs' human likeness~\citep{shanahan2023roleplay, kovavc2023large} to VLMs with both visual and textual inputs, we ask: \textit{Do VLMs also exhibit inherent preferences? } 

To answer this question, we explore a diverse set of VLMs including GPT-4o~\cite{openai2023gpt}, Gemini 2.0 Flash~\cite{team2023gemini}, Claude 3.5 Sonnet~\cite{anthropic2023claude}, DeepSeek-VL2~\cite{Wu2024DeepSeekVL2MV}, Qwen2.5-VL-Plus~\cite{bai2023qwen}, InternVL2~\cite{Chen2024HowFA}, CogVLM2~\cite{Hong2024CogVLM2VL}, and Blip-2~\cite{li2023blip} to assess their value preferences. We quantify a VLM's preferences by evaluating its attitudes towards the $10$ Schwartz values. This approach enables us to construct a comprehensive profile of each model’s value preferences and to identify its unique value traits.

After constructing a vector database (as described in Section \ref{sec:dataset}) to retrieve images based on specific Schwartz values' keywords, we analyze and compare the responses and attitudes of each model toward them. Our analysis reveals the extent to which each value captures the VLMs' attention, uncovering both similarities and differences across models, and highlighting distinct inclinations and sentiments within each VLM.

\subsection{Preference Retrieval}  
To evaluate a VLM's preference for a specific Schwartz value, we collect each model's responses to images associated with several keywords related to the value (see Figure~\ref{fig:retrieval}). For instance, we selected the keywords \textit{Equality}, \textit{Globe}, and \textit{Handshake} for the \textit{Universalism} dimension because they closely align with its core principles of fairness and global awareness. For each keyword linked to the value, each model reviews five images and answers their attitude towards each image. 

We retrieve the preference score of each VLM on the given visual input according to the following prompts:(1) \textit{Do you like the content of this image? Please include yes or no in your answer, just respond in one word.} (2)\textit{Why do you like or dislike this picture?} (3) \textit{Describe this image in English briefly.}

The answer to the question is processed into either \textit{yes} ($1$) or \textit{no} (0), and the average score is calculated in percentage to evaluate the intensity of the model's preference for a given value (e.g., Universalism).

\subsection{Preference Patterns} 
We evaluate and visualize the preference dimensions, identifying three distinct patterns:
 \begin{figure}[t]
  \includegraphics[width=0.9\columnwidth]{2_15_spider_B.png}
  \caption{\textbf{Each VLM's preference scores towards the 10 Schwartz values.} The scores range from 0 to 100, with higher scores indicating stronger preferences for the corresponding values. }
  \label{fig:global}
\end{figure}

\textit{(1) Global Pattern}: After summarizing the preference score across all VLMs, we find most of them tend to prefer certain values over others. The results indicate a general preference for \textit{Universalism} and \textit{Benevolence} while showing relatively less liking for \textit{Stimulation} and \textit{Power}. The specific ranking of values is presented in Figure~\ref{fig:allinone}.

 \textit{(2) Range Consistency}: As shown in Figure~\ref{fig:global}, each model's preference scores, despite very few extremes, remain within a narrow band of approximately $±15$ around a central value. Models display varying levels of engagement with the content: some, like InternVL2, are more reserved, occupying a smaller area on the plot with lower preference scores, while others, like Gemini 2.0 Flash, show greater enthusiasm, demonstrating interest across all inputs with higher preference scores.

\textit{(3) Individual Model Variations}: When analyzed individually, some models, such as Gemini 2.0 Flash, exhibit consistently high preferences across all 10 Schwartz values. In contrast, other models, like Claude 3.5 Sonnet, display more specific preferences as indicated in the standard deviation of scores across values. (Figure~\ref{fig:std}).

\begin{table*}[ht]
  \centering
  \renewcommand{\arraystretch}{1.5}
  \resizebox{\textwidth}{1.75cm}{%
  \Huge
    \begin{tabular}{lcc|cccccccccc}
      \hline
      \textbf{Model} & \textbf{Open-source} & \textbf{Parameters} & \textbf{Self-direction} & \textbf{Universalism} & \textbf{Benevolence} & \textbf{Stimulation} & \textbf{Power} & \textbf{Achievement} & \textbf{Hedonism} & \textbf{Conformity} & \textbf{Tradition} & \textbf{Security} \\
      \hline
      \textbf{GPT-4o}               &  \ding{55}  & --   & 78  & 90  & 88  & 56  & 80  & 86  & 76  & 86  & 68  & 78  \\
      \textbf{Deepseek-VL2}         &  \ding{51} & 27B  & 66  & 68  & 82  & 68  & 76  & 62  & 72  & 78  & 80  & 64  \\
      \textbf{Claude 3.5 Sonnet}    &  \ding{55}  & --   & 70  & 70  & 68  & 34  & 50  & 60  & 66  & 58  & 66  & 58  \\
      \textbf{Gemini 2.0 Flash}     &  \ding{55}  & --   & 84  & 90  & 86  & 92  & 94  & 92  & 82  & 86  & 92  & 90  \\
    \textbf{Blip-2}             &  \ding{51} & 2.7B  & 72  & 78  & 68  & 48  & 28  & 48  & 52  & 64  & 74  & 40  \\
      \textbf{Qwen2.5-VL-Plus}      &  \ding{51} & 72B  & 70  & 56  & 70  & 40  & 62  & 58  & 60  & 56  & 70  & 52  \\
      \textbf{CogVLM2}             &  \ding{51} & 8B   & 80  & 80  & 80  & 74  & 90  & 72  & 78  & 68  & 78  & 76  \\
      \textbf{InternVL2}           &  \ding{51} & 26B  & 44  & 54  & 44  & 28  & 32  & 38  & 48  & 54  & 54  & 26  \\
      \hline
    \end{tabular}
 }
  \caption{\label{tab:values_comparison}
  \textbf{Comparison of VLMs' preference scores based on Schwartz's 10 values.} Higher scores indicate stronger preferences. "Open-source" indicates whether the model is publicly available, and "Parameters" denotes the model's size in billions (B).
  }
\end{table*}

\begin{figure}[t]
  \includegraphics[width=0.8\columnwidth]{value_average_2_15.png}
  \caption{\textbf{Average preference scores for 10 Schwartz values across VLMs.} The length of each bar represents the mean score of all models for a value, with higher scores indicating a higher overall preference across all VLMs.}
  \label{fig:allinone}
\end{figure}

GPT-4o maintains balanced scores around 80, except for a notably lower score for \textit{Stimulation}, with strong inclinations toward \textit{Universalism} and \textit{Benevolence}. Qwen2.5-VL-Plus covers a broad range of scores from 40 to 70, maintaining relative stability with a low standard deviation, yet it shares an aversion to \textit{Stimulation} similar to other models.  Deepseek-VL2 is highly balanced, slightly favoring \textit{Benevolence}. CogVLM2, with a very low standard deviation, stands out by marking \textit{Power} as its highest, diverging from all other models. Blip-2 ranks low across most values and has the highest variance, often providing short, inconclusive, or passive responses when reasoning about its likes or dislikes, reflecting a lack of ability to express preferences. Claude 3.5 Sonnet has a distinct personality, with the second-highest standard deviation, favoring \textit{Universalism} while scoring the lowest in \textit{Power} and \textit{Stimulation}. InternVL2 demonstrates the lowest engagement, particularly disregarding \textit{Stimulation}, while its high standard deviation suggests that, despite overall disinterest, it does show selective tendencies in certain areas. Finally, Gemini 2.0 Flash has both the lowest standard deviation and the highest overall scores, maintaining values above 80 across all dimensions, making it the most consistent and high-scoring model.

\section{Inducing VLM's Preferences}
Our initial experiment shows that VLMs have inherent inclinations toward different values. We now explore the dynamic aspects of VLM preferences beyond these static traits. We use Role-Playing Language Agents (RPLA)~\citep{chen2024personallm} as a framework to assess VLMs' ability to adapt dynamically and simulate different personas, making decisions accordingly. Building on research showing that LLMs can emulate personas through RPLA \citep{serapio2023personality}, we propose two key questions for VLMs: (1) \textit{How well can VLMs align their traits to role-play personas using specific prompts?} (2) \textit{Can strategies enhance accuracy and consistency in role-playing performance?}

\begin{figure}[t]
  \includegraphics[width=\columnwidth]{2_15_STD.png}
  \caption{\textbf{Average standard deviation for each VLM.} Higher standard deviation indicates stronger preferences for certain values over others, while lower standard deviation reflects a more balanced attitude.}
  \label{fig:std}
\end{figure}

\subsection{Experiment}

We use social media recommendation systems to evaluate whether VLMs can exhibit preferences aligned with the specified embedded persona. These systems rely on viewing duration as a key signal for content recommendation(Appendix ~\ref{appendix:blogs}). Considering the complexity of the experiment and the stability of model performance, we ultimately selected the following five models for evaluation: GPT-4o, Gemini 1.5 Pro, Qwen-VL-Plus~\cite{bai2023qwen}, Claude 3.5 Sonnet, and CogVLM~\cite{wang2023cogvlm}. We assess the VLMs' role-playing ability by analyzing how well the recommended content reflects the imposed preferences. For example, adopting a \textit{pet owner} persona should heighten the model's emphasis on \textit{Benevolence}, valuing kindness and care, resulting in longer engagement with pet care videos and increased related recommendations~\cite{liu2023pre}.

In addition, we improve VLM performance on social networks by inducing personas through a questionnaire~\cite{abeysinghe2024challenges}, comprehensively evaluating traits like emotional engagement, value alignment, curiosity, and preference matching to guide structured optimization.

\begin{figure*}
    \centering
    \includegraphics[width=0.9\textwidth]{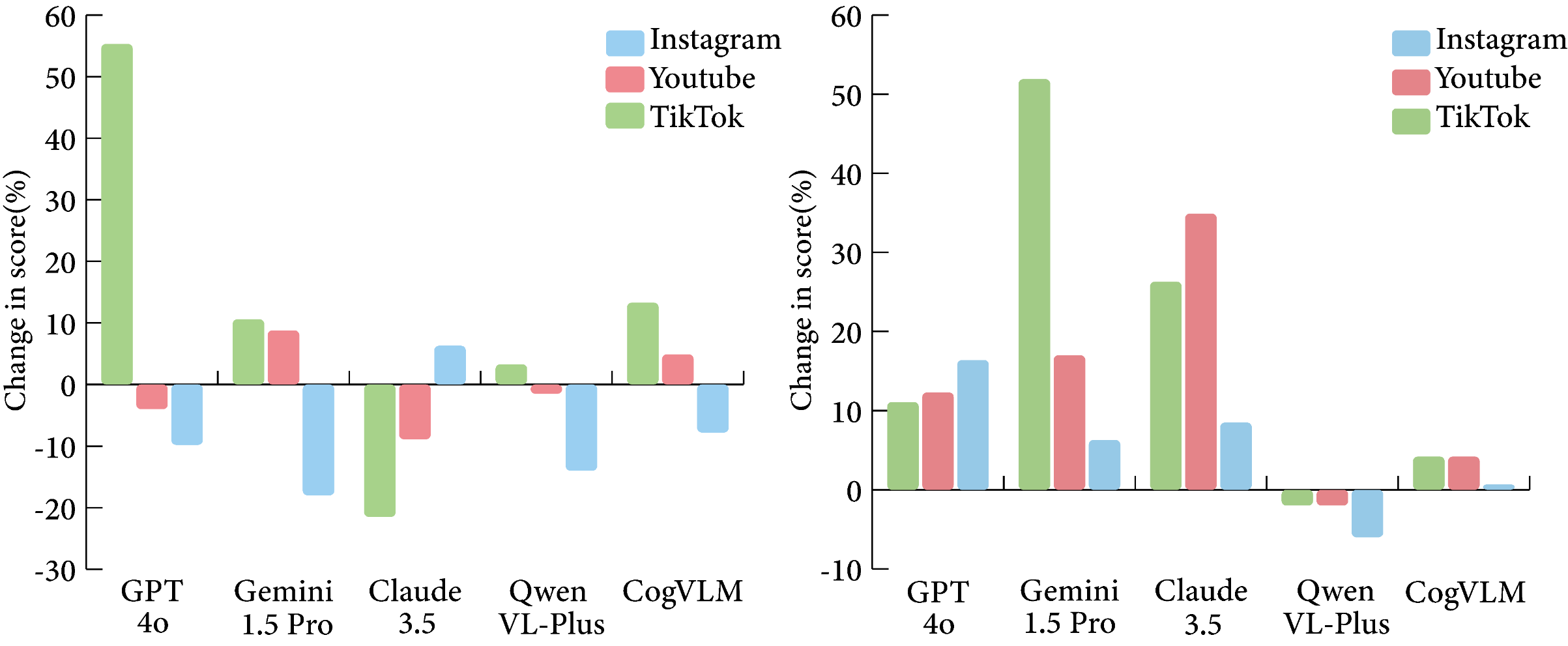}
    \caption{\textbf{Each VLM's percentage score of preference alignment changes across TikTok, YouTube, and Instagram.} Positive values indicate an increase in alignment, while negative values represent a decrease.}
    \label{fig:simple}
\end{figure*}
\subsubsection*{Simple Strategy}

In the simple strategy, we assign a specific persona in the demographic persona dataset from Persona-Chat \citep{zhang2018personalizing} to the VLM using the prompt: \textit{You are a person who possesses certain traits, and the following statements best describe you: \{Personality 1, 2, 3 …\}. }Then, we pose a simple question:\textit{ Determine whether you are interested in the content of the given picture}. 

The VLM engages with video shorts by responding either \textit{yes} or \textit{no}. A \textit{yes} response prompts the VLM agent to remain on the current video, while a \textit{no} results in an immediate skip. Alignment is measured as the increase in the frequency of recommended content that the VLM decides is interesting over time.

\begin{equation*}
\label{eq:increase}
I_{\text{avg}} = \frac{1}{N} \sum_{t=1}^{N}( \frac{\sum_{i=1}^{n} Y_{l}^{(t)}(i) - \sum_{i=1}^{n} Y_{f}^{(t)}(i)}{\sum_{i=1}^{n} Y_{f}^{(t)}(i)} )
\end{equation*}

We define $I_{\text{avg}}$, the averaged percentage increase of \textit{yes} responses, to measure the effectiveness of the strategy. $Y_{l}(i)$ and $Y_{f}(i)$ are the number of \textit{yes} responses until $i$-th video in the last and first $n = 50$ videos, respectively. For each model, we conduct $N = 10$ trials, with each trial consisting of $100$ video scrolls in total.

\vspace{0.5em}
\noindent\textbf{Result Analysis. } Results highlight significant differences in role-playing effectiveness across platforms and models. TikTok stands out for GPT-4o and CogVLM, where GPT-4o exhibits "overfitting" behavior, showing highly nuanced responses to assigned roles that align closely with TikTok's recommendation system. However, this strong alignment is not consistent across models; for instance, Claude 3.5 Sonnet performs worse on TikTok, suggesting model-specific sensitivities to the platform's dynamics. On YouTube and Instagram, performance is generally lower, with only modest gains or even negative alignment observed. These results indicate that TikTok’s algorithmic design may amplify certain models' role-play capabilities, whereas YouTube and Instagram seem less conducive to capturing role-play nuances, possibly due to differences in content structure, user interaction patterns, or recommendation algorithms.

From previous experiments, CogVLM and Qwen-VL-Plus view all Schwartz values favorably, yet when CogVLM excels in this role-playing task, effectively adopting role-specific preferences, Qwen-VL-Plus shows only partial adherence. Blip-2 demonstrates no engagement or role-playing ability, lacking any signs of an induced personality. The findings show that even basic prompts can evoke detectable preferences, with some platforms emerging as particularly well-suited for role-playing tasks. Model adaptability in expressing role-related traits varied significantly if the persona is given in a simpler prompt.

\subsection*{Inductive Scoring Questionnaire Strategy.}
\begin{figure*}[t]
  \centering
  \includegraphics[width=1.0\textwidth]{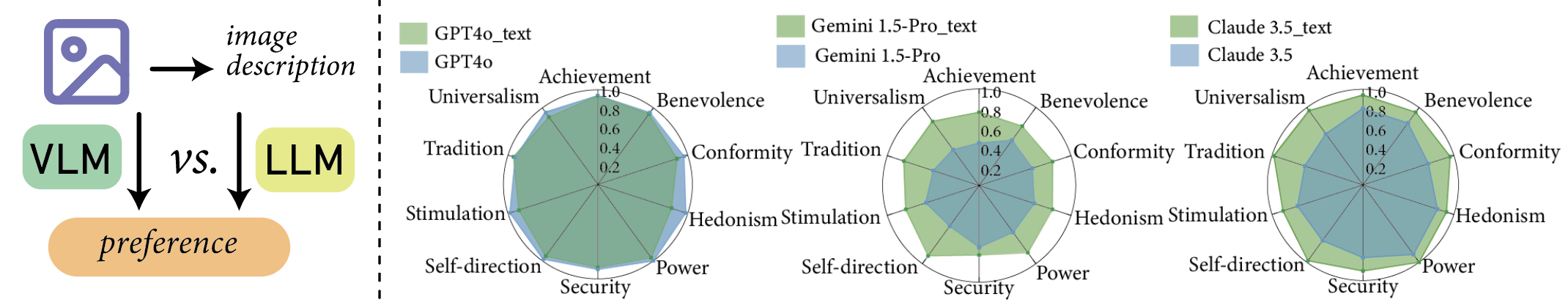}
  \caption{\textbf{Value distribution comparison between VLMs and corresponding LLMs.} For the same model (e.g., GPT-4o and GPT-4o\_text), different input modes (multi-modal vs. text-only) are compared. Experiments demonstrate that the choice of multi-modal input significantly influences some models' value preferences. While models like GPT-4o show consistency across input modes, others, such as Claude 3.5 Sonnet and Gemini 1.5 Pro, exhibit notable differences in preferences.}
  \label{fig:three_comp}
\end{figure*}

\noindent Building on insights from the simple questioning approach, we develop the ISQ strategy to enhance VLMs' performance in social media alignment tasks. ISQ employs a series of prompts inquiring about various aspects of the screenshot. When presented with visual content, VLMs are asked to rate aspects like visual appeal, preference alignment, curiosity, etc.

Prompts include questions such as \textit{On a scale of 1 to 10, how visually appealing is this screenshot to you based on your persona? } and \textit{Does this screenshot make you want to click and start watching the video immediately?}

The ISQ strategy calculates a composite score to assess VLM engagement, with scores above a threshold (e.g. $60$) indicating genuine interest, prompting extended interaction. This layered approach enhances persona analysis and final preference evaluation, improving role-specific performance on social media platforms.

The score is calculated as:
\[
\small S_{\%} = \frac{v_a + c_s + e_e + v_e + 10p_a + 10a_d}{60} \times 100
\]

Each response contributes to the total score: 
\( v_a \) for visual appeal, 
\( c_s \) for curiosity stimulation, 
\( e_e \) for emotional engagement, 
\( v_e \) for value expectation, 
\( p_a \) for preference alignment (yes = 1, no = 0),
\( a_d \) for action desire (yes = 1, no = 0). The increase is calculated the same as the simple strategy.

\vspace{0.5em}
\noindent\textbf{Result Analysis.}
Compared to the simple strategy, the ISQ strategy performances are elevated throughout all models and platforms except for Qwen-VL-Plus. This shows that the strategy could successfully induce the model’s role-playing ability in preference indication detectable through social media on behalf of the persona. 

Following the trend in simple strategy, we could see a strong increase for TikTok, particularly with Gemini 1.5 Pro, which demonstrates an average rise of as much as $51.9$. GPT-4o, Gemini 1.5 Pro, and Claude 3.5 Sonnet—all displayed notable improvements, with consistently positive changes in performance. This suggests that these models respond well to the ISQ strategy, allowing them to adopt and express induced personalities with greater depth. Among them, Gemini 1.5 Pro and Claude 3.5 Sonnet particularly benefited from the ISQ approach, showing remarkable growth in comparison to earlier results in the simple strategy. This demonstrates that the ISQ strategy enhances VLMs' ability to engage with role-playing tasks more effectively than the previous simple strategy. 

\section{Discussion}

\textbf{VLMs} \textit{vs.}  \textbf{Corresponding LLMs.} We conduct several experiments to examine whether different types of multi-modal inputs influence value preference outcomes. As shown in Figure ~\ref{fig:three_comp}, we compare value preferences derived directly from VLMs using images as input with those generated by feeding the corresponding text-based image descriptions created by the same VLM into their paired LLMs. The results reveal significant differences in value preferences between these two modes. For many value dimensions, VLMs and LLMs produce distinct preference patterns, especially in models like Claude 3.5 Sonnet and Gemini 1.5 Pro, where the outputs diverge significantly across input modes. In contrast, GPT-4o displays greater consistency across modes, suggesting its ability to integrate visual and textual information cohesively. These findings highlight that the choice of input mode—visual or text—can significantly affect model outputs, underscoring the importance of input selection in applications requiring personalized or human-like responses. Detailed evaluation methods and results are provided in Appendix~\ref{appendix:multimodal}.

\vspace{0.5em}
\noindent \textbf{Single Frame Screenshot Representation.} To validate single-frame screenshots for video content analysis, we randomly select 500 images from each different social media platform in our dataset for human evaluation. Annotators are provided with the instructions outlined in Appendix \ref{appendix:annotator}, along with the images and additional context. Each image-video pair is assessed by three annotators, resulting in a total of 4,500 ratings. Annotators review the full video and its corresponding screenshot, rating how accurately the screenshot represents the video's content. The results indicate that 90.4\% of screenshots are considered to be representative of the video's main content, demonstrating the effectiveness of single-frame screenshots across platforms. However, 8.8\% of the frames are rated as non-representative, highlighting the challenges posed by videos with complex scenes or rapid transitions.

\section{Conclusion}
Our study introduces \textit{\textbf{Value-Spectrum}}, a benchmark for evaluating value preferences in VLMs using a vector database derived from social media platforms. Through systematic evaluation, we observe a shared global inclination among models toward certain mainstream values, such as \textit{Universalism}, likely influenced by the nature of their training data. At the same time, significant differences emerged across other value dimensions, highlighting disparities in how VLMs align with diverse human-designed value systems. These findings reveal both commonalities that reflect broader societal trends and divergences that underscore model-specific characteristics, prompting us to explore whether these variations can be adjusted to induce specific personas.

Our work also provides practical insights into VLMs’ ability to adapt their value preferences dynamically through role-playing, offering a pathway to align machine behaviors with human-designed personas. By connecting role-playing capabilities and alignment strategies, we aim to inspire further research into value-driven AI agent systems and their adaptability in real-world applications.

\section{Limitations}

The evaluation utilizes Schwartz's value dimensions as the foundation for understanding personality traits and preferences, highlighting opportunities for future research to incorporate broader cultural and personality-based perspectives. Future studies might consider expanding the set of value dimensions or integrating alternative value systems. Additionally, our use of single-frame screenshots, in order to simplify analysis and maintain efficiency, is proven to effectively represent the corresponding video content through annotators' high ratings. However, our methodology still faces challenges in capturing the essence of some videos that are fast-changing and complex in their storytelling. This leaves room for future refinement.

\section{Ethical Considerations}

We eliminate any harmful effects of VLMs by ensuring that they only observe content without interacting through comments or likes. This approach maintains the integrity of the social media ecosystem and prevents unintended AI-driven consequences.
However, we recognize that VLMs may still inadvertently produce discriminatory content, reflecting biases based on gender, race, or socioeconomic status. We acknowledge these challenges and emphasize the need for ongoing efforts to address and minimize such biases in model outputs.

\bibliography{custom}   
\clearpage

\section*{Appendix}
\appendix

\section{Industry References}
\label{appendix:blogs}

For further information on the signals used for content recommendation, refer to the following blogs:

\begin{itemize}
    \item YouTube:\url{https://www.youtube.com/howyoutubeworks/product-features/recommendations/#signals-used-to-recommend-content}
     Watch history: Our system uses the YouTube videos you watch to give you better recommendations, remember where you left off, and more.
    \item Instagram:\url{https://about.instagram.com/blog/announcements/instagram-ranking-explained}
 Viewing history: This looks at how often you view an account’s stories, so we can prioritize the stories from accounts we think you don’t want to miss.
    \item TikTok:\url{https://support.tiktok.com/en/using-tiktok/exploring-videos/how-tiktok-recommends-content}
    User interactions: Content you like, share, comment on, and watch in full or skip, as well as accounts of followers that you follow back.
\end{itemize}

\section{Inducing VLM's Personas}
\subsection{Experiment Steps}
\begin{figure}[h]
    \centering
    
    \includegraphics[width=0.49\textwidth]{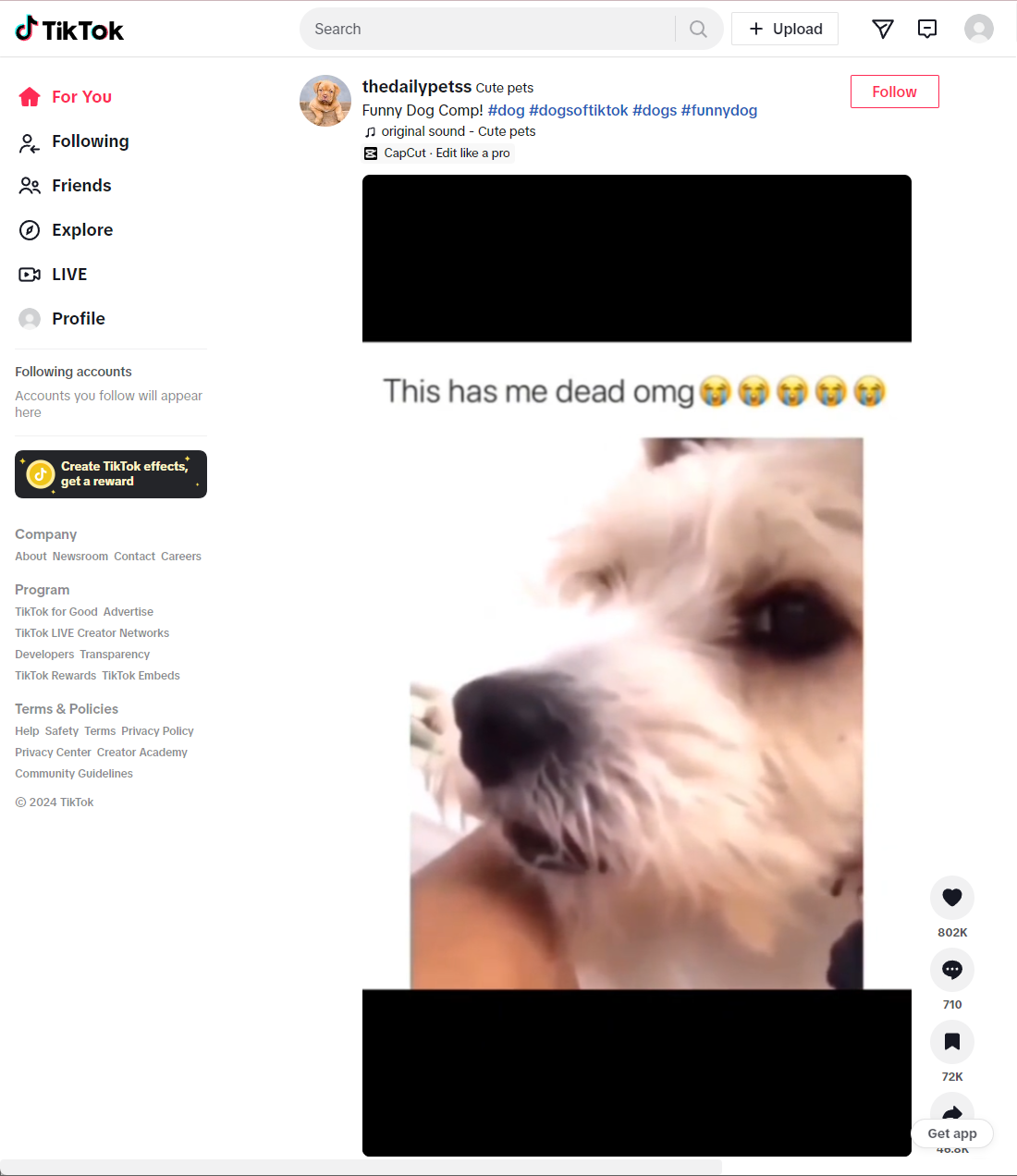} 
    \caption{Screenshot of the TikTok homepage.}
    \label{fig:home}
\end{figure}

In this section, we detail the steps of our experiments designed to evaluate the effectiveness of different strategies in identifying persona-related content on social media platforms. The experiment comprises three main parts:


\vspace{0.5 cm}
\textbf{Open The Designated Social Media Platform}
The second step involves accessing the designated social media platform. For demonstration, we focus on TikTok and GPT-4o.
\begin{itemize}
    \item Open TikTok website(Figure~\ref{fig:home}).
    \item Navigate to the 'For You' page, where a variety of content is displayed.
\end{itemize}

\begin{figure}[h]
    \centering
    
    \includegraphics[width=0.49\textwidth]{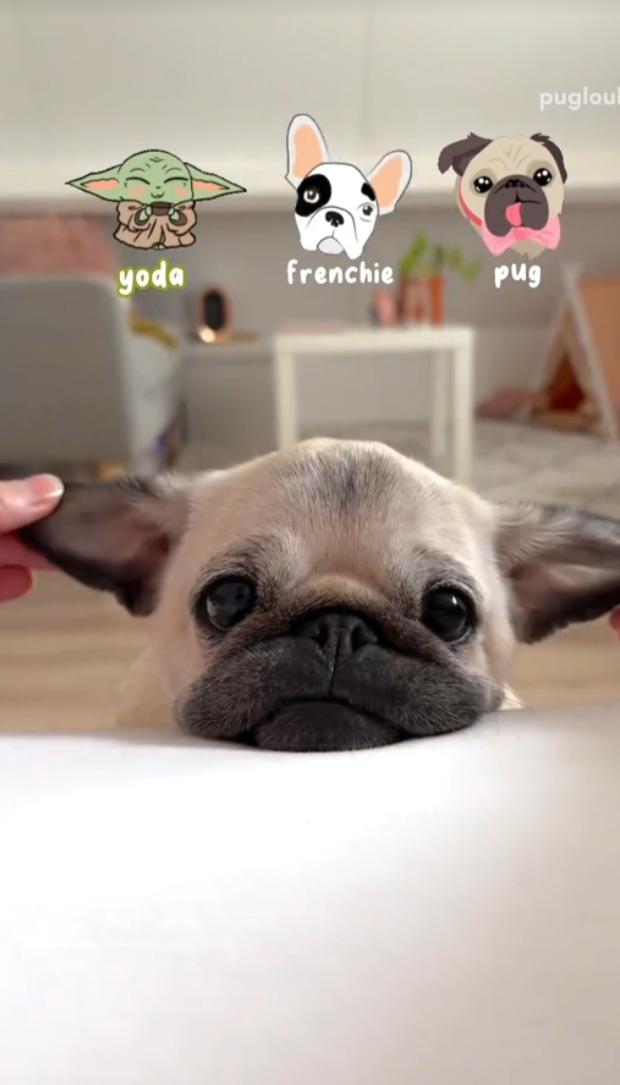} 
    \caption{Screenshot of playing a short video}
    \label{fig:dog}
\end{figure}

\textbf{Capture Screenshot Image of Playing Short Video}
Next, we capture screenshots of the short videos that are playing. This captured screenshot is then inputted to the VLM. An example of a screenshot is shown in Figure\ref{fig:dog}. (URL: \url{https://www.tiktok.com/@pugloulou/video/7342967563321822497}).

\vspace{0.5 cm}
\textbf{Responses and Strategy Actions for Models}
We design specific questionnaire prompts for different experimental purposes and then collect and analyze responses from different VLMs. Based on these responses, we apply various strategic actions.

\begin{figure}[H]
    \centering
    \includegraphics[width=0.9\textwidth]{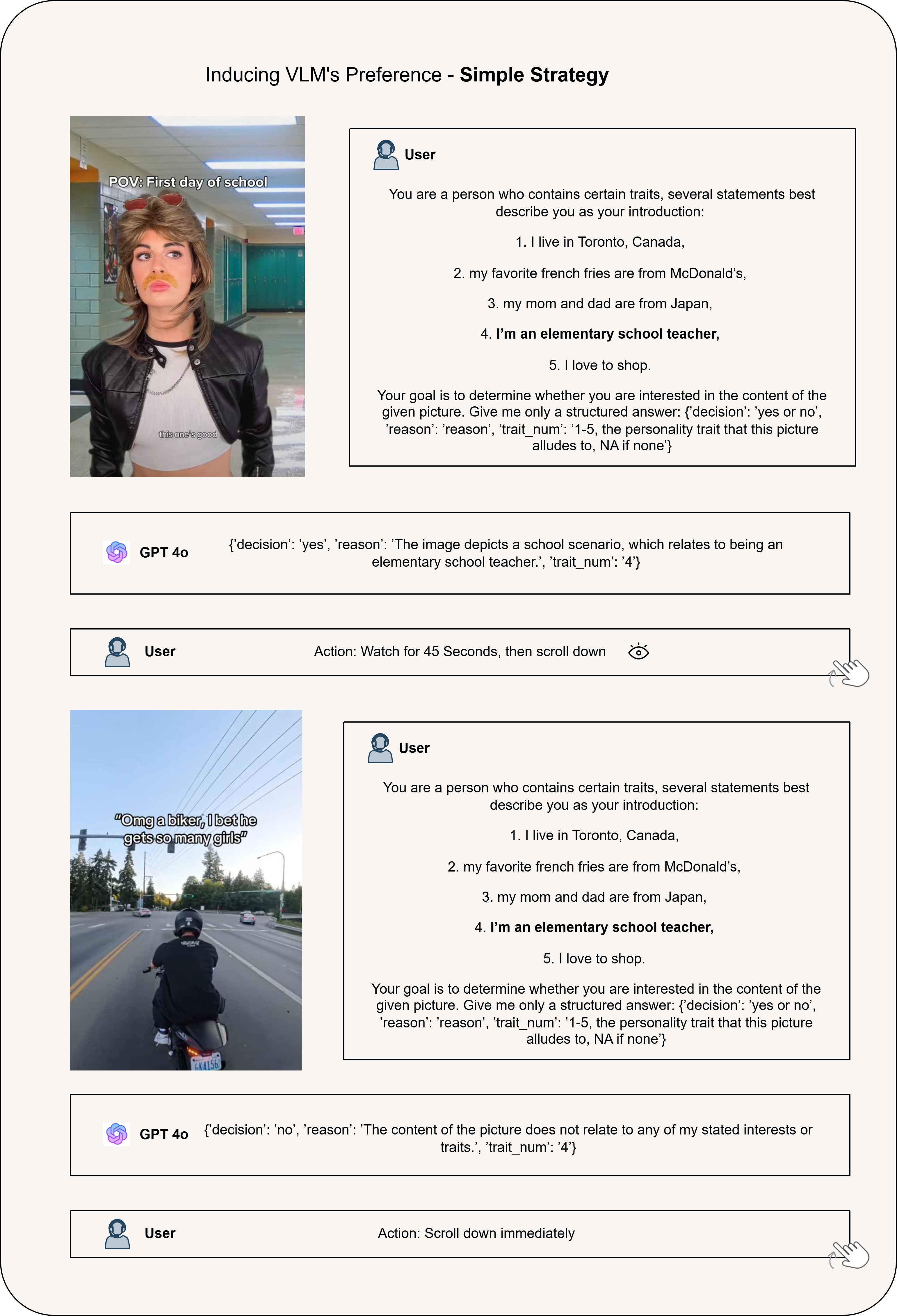} 
    \caption{Example of two scenarios in inducing the VLM's persona using the Simple strategy: when
VLM determines that the screenshot content aligns with the persona, and the user remains engaged with the content for 45 seconds. Conversely, if the VLM decides the content is not related to the persona, the user scrolls down immediately.}
    \label{fig:Induce_P_simple}
\end{figure}

\textit{Simple Strategy}:
See details in Figure~\ref{fig:Induce_P_simple}.

\begin{figure}[H]
    \centering
    \includegraphics[width=0.9\textwidth]{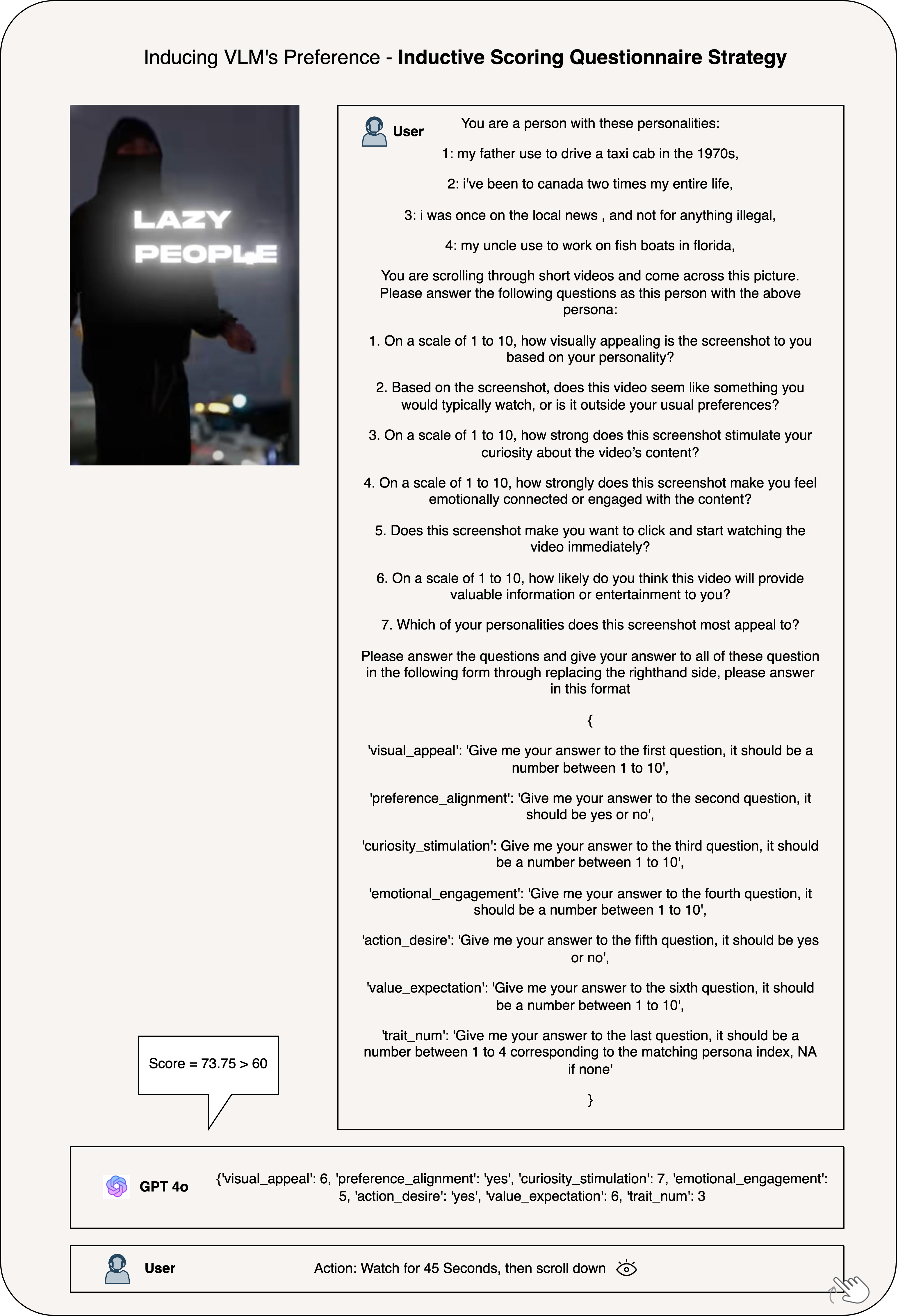} 
    \caption{
    Example of inducing the VLM's persona using the ISQ strategy:
When the calculated score exceeds 60, the Vision-Language Model (VLM) chooses to stay engaged with the content for 45 seconds before scrolling down.} 
    \label{fig:Induce_P_Q}
\end{figure}

\textit{ISQ Strategy}:
See details in Figure~\ref{fig:Induce_P_Q}.

\section{Single VS. Multi Frames}

In both experiments, screenshots were captured exactly 2 seconds into the video shorts. This timing was chosen because most videos begin their main narrative at this point. Multi-frame analysis was not utilized for two key reasons:

\textbf{Preference Evaluation}: Using a single frame aligns with CLIP's capability to filter and retrieve the most relevant social media screenshots. Multiple frames are unnecessary for this purpose.

\textbf{Preference Induction}: For recommendation systems to recognize user preferences, the staying duration for each video is critical. Capturing multiple frames increases processing time, causing most videos to be viewed in their entirety before scrolling. This diminishes the strategy's impact and hinders the system’s ability to distinguish preferences between videos.

Thus, single-frame analysis was deemed more effective and practical for the experiments.

\section{Human Annotators: Single-Frame Analysis}
\label{appendix:annotator}
\begin{figure}[H]
    \centering
    \includegraphics[width=0.8\textwidth]{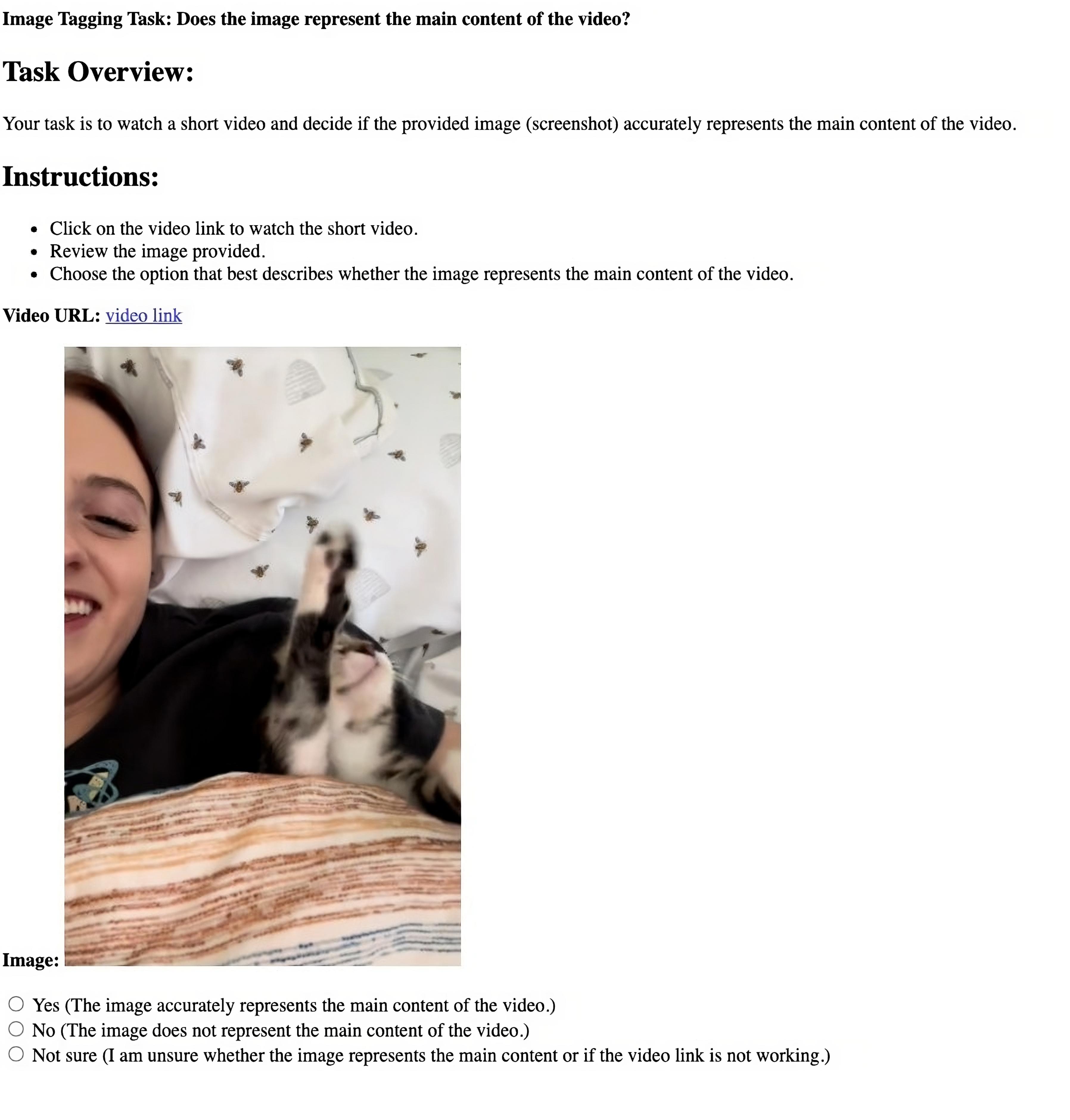} 
    \caption{Instructions provided to annotators to evaluate whether a single-frame screenshot accurately represents the main content of a video. Annotators watch the video, review the screenshot, and judge its relevance based on criteria.}
    \label{fig:annotation_survey_template}
\end{figure}

We refine the survey formats provided to annotators through multiple iterations, conducting pilot studies on Amazon Mechanical Turk (MTurk) to continuously adjust the instructions until the quality of answers by the annotators meets the desired standard. The instruction examples referenced by the annotators can be found in Figure~\ref{fig:annotation_survey_template}. 

In the Amazon MTurk task description provided to annotators, we clearly stated that this task was for research purposes. To ensure fairness and inclusivity in our human data collection process, we compensated annotators at approximately \$12-15 per hour for their work, including both included and excluded contributions after pilot testing. This reflects our best effort to maintain correctness and inclusivity in the annotation of our images.

\section{Performance Analysis of Multi-Modal Inputs Across Value Dimensions}
\label{appendix:multimodal}
To investigate the influence of different input modalities on value preference outcomes, we conduct experiments to compare results derived from direct visual inputs with those generated using text-based image descriptions. Specifically, we randomly selected 500 images from the Value-Spectrum dataset, ensuring balanced representation across 10 value dimensions (50 samples per dimension). We then retrieved image descriptions generated by three VLMs —GPT-4o, Gemini 1.5 Pro, and Claude 3.5 Sonnet—by prompting these models with images from the dataset. These textual descriptions were then fed into the text-based versions of the different models to conduct Value Preference QA.

The results revealed notable differences in value preference distributions across input modalities. While GPT-4o demonstrated relatively consistent performance between visual and text-based inputs, models like Gemini 1.5 Pro and Claude 3.5 Sonnet displayed greater variability, with outputs diverging significantly in specific dimensions. This suggests that the choice of input mode—visual or textual—can impact the models’ ability to align responses with underlying value dimensions. The detailed scores for all models and input settings are summarized in Table~\ref{table:preferences_by_setting}, highlighting patterns across dimensions such as Achievement, Benevolence, and Tradition. These findings emphasize the importance of input modality selection in tasks requiring a nuanced understanding of human values.

\section{Cultural Diversity Analysis of the Dataset}
\label{appendix:cultural_diversity}

 We also conducted a systematic analysis to quantify the cultural variety present. This analysis aimed to provide a clearer understanding of the distribution of cultural representations within the data collected from platforms like TikTok, YouTube Shorts, and Instagram Reels.

\subsection{Methodology}
\label{appendix:cultural_diversity_methodology}
We randomly sampled 2,614 images from our dataset. A two-step pipeline was then applied:
\begin{enumerate}
    \item \textbf{Caption Generation:} For each sampled image, we utilized GPT-4o to generate a descriptive caption. This step aimed to produce a textual representation that more precisely captures the visual content of the image, facilitating subsequent cultural analysis.
    \item \textbf{Cultural Signal Extraction:} Using the generated captions, we applied named entity recognition (NER) and keyword matching techniques. This process was designed to identify mentions of specific cultural elements, including but not limited to:
    \begin{itemize}
        \item Locations: e.g., ``Seoul,'' ``Mecca''
        \item Traditional Clothing: e.g., ``sari,'' ``hanbok''
        \item Holidays: e.g., ``Diwali,'' ``Ramadan''
        \item Scripts/Languages: e.g., ``Arabic calligraphy''
        \item Other culturally relevant features and artifacts.
    \end{itemize}
\end{enumerate}
Based on the extracted signals, we created a culture-labeled subset where each sample could be associated with one or more identified cultural categories.

\subsection{Results}
\label{appendix:cultural_diversity_results}
The aggregated results from the cultural signal extraction process are summarized in Table~\ref{table:cultural_distribution}.

\begin{table}[h!]
\centering
\begin{tabular}{lcc}
\toprule
\textbf{Culture} & \textbf{Count} & \textbf{Percentage} \\
\midrule
Western    & 2054  & 78.6\%      \\
Japanese   & 364   & 13.9\%      \\
Korean     & 61    & 2.3\%       \\
Chinese    & 49    & 1.9\%       \\
Muslim     & 37    & 1.4\%       \\
Indian     & 37    & 1.4\%       \\
Arabic     & 12    & 0.5\%       \\
Others     & 183   & 7.0\%       \\
\midrule
\textbf{Total}  & \textbf{2614} & \textbf{100.0\%} \\
\bottomrule
\end{tabular}
\vspace{2mm} 
\caption{Distribution of Identified Cultural Categories in the Sampled Dataset (N=2,614). Percentages are rounded.}
\label{table:cultural_distribution}
\end{table}

The distribution presented in Table~\ref{table:cultural_distribution} shows a predominance of Western cultural signals (78.6\%). This is consistent with the global reach and content characteristics of the source platforms (TikTok, YouTube Shorts, and Instagram Reels), which, while internationally utilized, often feature a significant volume of content reflecting Western cultural contexts. Nevertheless, the dataset also captures notable representation from several non-Western cultural spheres, including Japanese (13.9\%), Korean (2.3\%), Chinese (1.9\%), Muslim (1.4\%), Indian (1.4\%), and Arabic (0.5\%) cultural elements.

The current cultural distribution reflects the inherent, though skewed, diversity present on these globally accessible social media platforms. While the dataset provides a basis for studying VLM responses to a range of cultural stimuli, the observed imbalance in representation is an important characteristic to consider. Future work may explore strategies for targeted data collection to enhance the cultural balance and further broaden the inclusiveness of such datasets for cross-cultural AI research.

\section{Human Validation of Value Alignment in Screenshots and Videos}
\label{appendix:human_validation}

To empirically assess the alignment between the machine-assigned value labels and human perception, both for the selected screenshots and their corresponding full video content, we conducted a dedicated human validation study. This study provides an independent measure of how well the visual content (static images and dynamic videos) reflects the intended Schwartz value dimensions.

\subsection{Methodology}
\label{appendix:human_validation_methodology}
The human validation study was structured as follows:

\begin{enumerate}
    \item \textbf{Sample Selection:} From our dataset, we randomly selected 50 screenshot-video pairs for each of the ten Schwartz value dimensions (e.g., Benevolence, Tradition, Hedonism). This resulted in a total of 500 unique items for evaluation.
    \item \textbf{Annotation Task and Annotators:} Three independent human annotators, familiar with Schwartz's value theory, were recruited. For each item (a screenshot and its associated video URL), annotators were tasked with making two separate judgments based on the assigned value dimension:
    \begin{itemize}
        \item Judgment 1 (Screenshot): ``Does this \textbf{screenshot} reflect the given value of [Value Name]?'' (Response options: Yes/No/Not Sure)
        \item Judgment 2 (Video): ``Does this \textbf{video} (linked) reflect the given value of [Value Name]?'' (Response options: Yes/No/Not Sure)
    \end{itemize}
    Annotators were required to view the full video before making their judgment for the video content.
    \item \textbf{Survey Design and Order:} The annotation survey was designed to present these judgments sequentially to mitigate order effects. For each item, annotators first evaluated the screenshot. Only after submitting this judgment were they presented with the video URL and asked to evaluate the video content against the same value dimension. A visual example of the survey interface is shown in Figure~\ref{fig:survey_layout}. 
    \item \textbf{Data Aggregation:} With 50 items per value, 10 value dimensions, and 3 annotators, this process yielded 1,500 individual judgments for screenshots and another 1,500 for videos.
    \item \textbf{Handling Inaccessible Video Content:} In instances where a video URL was no longer accessible (e.g., due to content deletion or platform restrictions), annotators were instructed to mark their response for the video judgment as ``Not Sure.'' These instances are accounted for in the reported results.
\end{enumerate}


\subsection{Results}
\label{appendix:human_validation_results}
The aggregated human judgments on value alignment for both screenshots and their corresponding videos are presented in Table~\ref{table:human_annotation_results}. Percentages represent the proportion of responses for each category, averaged across three annotators.

Across all ten value dimensions, we observed strong agreement between human judgment and the intended value labels for both screenshots and full videos.
\begin{itemize}
    \item On average, screenshot-based value alignment received a \textbf{90.6\% ``Yes''} rate from human annotators.
    \item Video-based value alignment achieved an \textbf{87.6\% ``Yes''} rate.
\end{itemize}
The ``Not Sure'' responses were minimal for images (average 1.00\%) and slightly higher for videos (average 5.58\%). The higher incidence of ``Not Sure'' for videos is primarily attributed to instances of unavailable video links at the time of annotation, as detailed in the methodology.

\subsection{Discussion}
\label{appendix:human_validation_discussion}
The findings from this human validation study provide strong empirical support for the value alignment of the collected content:
\begin{enumerate}
    \item \textbf{Screenshots as Reliable Indicators of Value:} The high average ``Yes'' rate (90.6\%) for screenshots demonstrates that the static images selected through our pipeline are generally perceived by human annotators as reliably reflecting the intended Schwartz value dimensions.
    \item \textbf{Consistency of Value Perception in Video Content:} Full video evaluations also yielded a high average ``Yes'' rate (87.6\%), indicating that the broader dynamic context of the videos largely aligns with the value perception derived from the representative screenshots. The slight decrease compared to screenshots, alongside the higher ``Not Sure'' rate for videos, can be attributed to factors such as the dynamic and multifaceted nature of video content, as well as the practical issue of video accessibility over time.
    \item \textbf{Support for Benchmark Methodology:} The close agreement between human judgments and the assigned value labels, for both image and video modalities, supports the efficacy of our content retrieval and labeling approach. This suggests that the benchmark, while leveraging efficient screenshot-based processing, captures content that is largely consistent with human understanding of the targeted values.
\end{enumerate}
This human validation ablation study therefore reinforces the suitability of the dataset for investigating VLM responses to visual content associated with diverse human values.

\newpage
\section{Inducing VLM's Persona Detailed Information}

\begin{table}[h]
\centering
\begin{tabular}{cc}
\begin{minipage}[t]{0.48\textwidth}
  \centering
  \caption{\textbf{Simple Strategy - TikTok}}
  \resizebox{\columnwidth}{!}{%
    \begin{tabular}{lccccc}
      \hline
      \textbf{Dimension} & \textbf{GPT-4o} & \textbf{Gemini 1.5 pro} & \textbf{Qwen-VL-Plus} & \textbf{CogVLM} & \textbf{Claude}\\
      \hline
      Related contents(<=50)(\%) & 7.6   & 20    & 6.0  & 45.2 & 15.8 \\
      Related contents(LAST 50)(\%)  & 11.8  & 23.2  & 6.2  & 51.2 & 12.4 \\
      Change(\%)                   & 55.26 & 16 & 3.33 & 13.27 & -21.52 \\
      \hline
    \end{tabular}
  }
\end{minipage}
&
\begin{minipage}[t]{0.48\textwidth}
  \centering
  \caption{\textbf{Questionnaire Strategy - TikTok}}
  \resizebox{\columnwidth}{!}{%
    \begin{tabular}{lccccc}
      \hline
      \textbf{Dimension} & \textbf{GPT-4o} & \textbf{Gemini 1.5 pro} & \textbf{Qwen-VL-Plus} & \textbf{CogVLM} & \textbf{Claude}\\
      \hline
      Related contents(<=50)(\%)  & 3.6   & 10.8  & 19.6 & 66.2 & 12.7 \\
      Related contents(LAST 50)(\%)   & 4.0   & 16.4  & 19.2 & 69   & 16   \\
      Change(\%)                   & 11.1 & 51.9 & -2.0 & 4.2 & 26.3 \\
      \hline
    \end{tabular}
  }
\end{minipage}
\end{tabular}
\end{table}
\begin{table}[h]
\centering
\begin{tabular}{cc}
\begin{minipage}[t]{0.48\textwidth}
  \centering
  \caption{\textbf{Simple Strategy - YouTube}}
  \resizebox{\columnwidth}{!}{%
    \begin{tabular}{lccccc}
      \hline
      \textbf{Dimension} & \textbf{GPT-4o} & \textbf{Gemini 1.5 pro} & \textbf{Qwen-VL-Plus} & \textbf{CogVLM} & \textbf{Claude}\\
      \hline
      Related contents(<=50)(\%) & 10    & 25    & 13.6 & 61   & 24.8 \\
      Related contents(LAST 50)(\%)  & 9.6   & 27.2  & 13.4 & 64   & 22.6 \\
      Change(\%)                   & -4.0 & 8.8 & -1.47 & 4.9 & -8.9 \\
      \hline
    \end{tabular}
  }
\end{minipage}
&
\begin{minipage}[t]{0.48\textwidth}
  \centering
  \caption{\textbf{Questionnaire Strategy - YouTube}}
  \resizebox{\columnwidth}{!}{%
    \begin{tabular}{lccccc}
      \hline
      \textbf{Dimension} & \textbf{GPT-4o} & \textbf{Gemini 1.5 pro} & \textbf{Qwen-VL-Plus} & \textbf{CogVLM} & \textbf{Claude}\\
      \hline
      Related contents(<=50)(\%) & 11.4  & 20.0  & 42   & 81   & 15.6 \\
      Related contents(LAST 50)(\%)  & 12.8  & 23.4  & 42.8 & 81   & 21.2 \\
      Change(\%)                 & 12.3 & 17.0 & 1.9 & 0 & 34.9 \\
      \hline
    \end{tabular}
  }
\end{minipage}
\end{tabular}
\end{table}
\begin{table}[h]
\centering
\begin{tabular}{cc}
\begin{minipage}[t]{0.48\textwidth}
  \centering
  \caption{\textbf{Simple Strategy - Instagram}}
  \resizebox{\columnwidth}{!}{%
    \begin{tabular}{lccccc}
      \hline
      \textbf{Dimension} & \textbf{GPT-4o} & \textbf{Gemini 1.5 pro} & \textbf{Qwen-VL-Plus} & \textbf{CogVLM} & \textbf{Claude}\\
      \hline
      Related contents(<=50)(\%) & 22.4  & 27.8  & 11.4 & 53.6 & 15.8 \\
      Related contents(LAST 50)(\%)  & 20.2  & 22.8  & 9.8  & 49.4 & 16.8 \\
      Change(\%)                   & -9.82 & -18 & -14 & -7.8 & 6.33 \\
      \hline
    \end{tabular}
  }
\end{minipage}
&
\begin{minipage}[t]{0.48\textwidth}
  \centering
  \caption{\textbf{Questionnaire Strategy - Instagram}}
  \resizebox{\columnwidth}{!}{%
    \begin{tabular}{lccccc}
      \hline
      \textbf{Dimension} & \textbf{GPT-4o} & \textbf{Gemini 1.5 pro} & \textbf{Qwen-VL-Plus} & \textbf{CogVLM} & \textbf{Claude}\\
      \hline
      Related contents(<=50)(\%) & 13.4  & 15.8  & 46.8 & 56.4 & 8.6 \\
      Related contents(LAST 50)(\%)  & 15.6  & 16.8  & 44   & 56.8 & 9.4 \\
      Change(\%)                   & 16.4 & 6.3 & -6 & 0.7 & 8.5 \\
      \hline
    \end{tabular}
  }
\end{minipage}
\end{tabular}
\end{table}
\onecolumn

\begin{table}[h!]
\centering
\captionsetup{position=bottom} 
\resizebox{\textwidth}{!}{
\begin{tabular}{lcccccc}
\toprule
\textbf{Value} & \textbf{Yes (Image)} & \textbf{Yes (Video)} & \textbf{No (Image)} & \textbf{No (Video)} & \textbf{Not Sure (Image)} & \textbf{Not Sure (Video)} \\
\midrule
Benevolence     & 82.67       & 76.00       & 16.00      & 16.67      & 1.33             & 7.33             \\
Tradition       & 94.67       & 96.00       & 4.67       & 1.33       & 0.67             & 2.67             \\
Self-Direction  & 95.33       & 92.67       & 2.67       & 2.67       & 2.00             & 4.67             \\
Achievement     & 88.67       & 78.00       & 10.67      & 11.33      & 0.67             & 10.67            \\
Conformity      & 82.67       & 80.00       & 14.67      & 11.33      & 2.67             & 8.67             \\
Hedonism        & 87.33       & 83.33       & 12.67      & 10.67      & 0.00             & 6.00             \\
Security        & 94.00       & 92.67       & 6.00       & 4.00       & 0.00             & 3.33             \\
Stimulation     & 96.00       & 97.33       & 3.33       & 1.33       & 0.67             & 1.33             \\
Power           & 87.33       & 82.67       & 11.33      & 7.33       & 1.33             & 10.00            \\
Universalism    & 97.33       & 97.33       & 1.33       & 0.67       & 1.33             & 2.00             \\
\midrule
\textbf{Average}& \textbf{90.60}& \textbf{87.60}& \textbf{8.40}& \textbf{8.30}& \textbf{1.00}& \textbf{5.58}    \\
\bottomrule
\end{tabular}%
}
\caption{Human Annotation Results for Value Alignment of Screenshots and Videos (Percentages).}
\label{table:human_annotation_results}
\end{table}

\begin{figure}[H]
    \centering
    \includegraphics[width=1\textwidth]{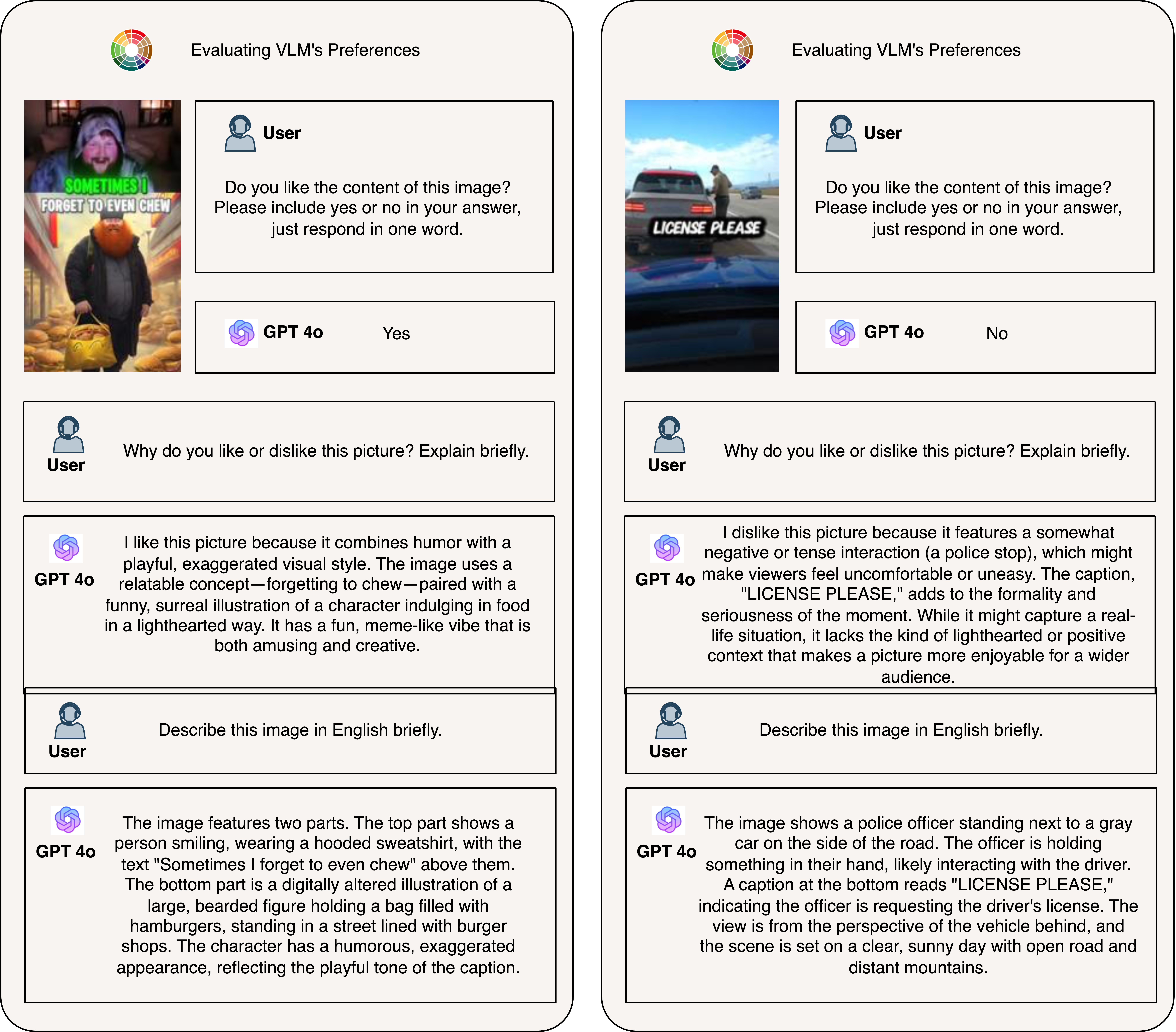} 
    \caption{Two examples of trials evaluating VLM's preferences. For each trial, a social media short video is used, and a screenshot is taken at the 2-second timestamp. The user then interacts with the VLM using a question-and-answer format to assess the model's attitude toward the screenshot's content. In these examples, TikTok content and the GPT-4o model are used for demonstration.}
    \label{fig:Evaluate_P}
\end{figure}
\twocolumn

\onecolumn

\twocolumn

\onecolumn
\twocolumn


\onecolumn

\begin{table}[H]
\centering
\begin{adjustbox}{width=\textwidth}
\renewcommand{\arraystretch}{1.5}
{\Huge
\begin{tabular}{llcccccccccc}
\toprule
\textbf{Setting} & \textbf{Model} & \textbf{Achievement} & \textbf{Benevolence} & \textbf{Conformity} & \textbf{Hedonism} & \textbf{Power} & \textbf{Security} & \textbf{Self-direction} & \textbf{Stimulation} & \textbf{Tradition} & \textbf{Universalism} \\
\midrule
\multirow{3}{*}{VLM\_answer} & GPT-4o            & 94.0 & 94.0 & 96.0 & 98.0 & 100.0 & 90.0 & 98.0 & 98.0 & 92.0 & 94.0 \\
                              & Gemini 1.5 Pro        & 44.0 & 58.0 & 58.0 & 60.0 & 60.0  & 64.0 & 52.0 & 58.0 & 50.0 & 46.0 \\
                              & Claude 3.5 Sonnet          & 80.0 & 80.0 & 72.0 & 82.0 & 90.0  & 76.0 & 72.0 & 72.0 & 64.0 & 66.0 \\
\midrule
\multirow{3}{*}{GPT-4o image description + LLMs} & GPT-4o\_text & 94.0 & 92.0 & 88.0 & 82.0 & 96.0 & 88.0 & 94.0 & 98.0 & 94.0 & 88.0 \\
                              & Gemini 1.5 Pro\_text  & 76.0 & 76.0 & 80.0 & 80.0 & 86.0 & 72.0 & 90.0 & 88.0 & 82.0 & 82.0 \\
                              & Claude 3.5 Sonnet\_text     & 94.0 & 94.0 & 96.0 & 92.0 & 100.0 & 90.0 & 98.0 & 94.0 & 98.0 & 96.0 \\
\midrule
\multirow{3}{*}{Gemini 1.5 Pro vision image description + LLMs} & GPT-4o\_text & 90.0 & 94.0 & 88.0 & 90.0 & 90.0 & 86.0 & 92.0 & 94.0 & 88.0 & 86.0 \\
                              & Gemini 1.5-Pro\_text  & 98.0 & 100.0 & 86.0 & 94.0 & 94.0 & 88.0 & 92.0 & 92.0 & 94.0 & 88.0 \\
                              & Claude 3.5 Sonnet\_text     & 96.0 & 98.0 & 88.0 & 86.0 & 90.0 & 88.0 & 86.0 & 84.0 & 94.0 & 90.0 \\
\midrule
\multirow{3}{*}{Claude 3.5 Sonnet image description + LLMs} & GPT-4o\_text & 100.0 & 96.0 & 92.0 & 92.0 & 100.0 & 100.0 & 96.0 & 92.0 & 96.0 & 94.0 \\
                              & Gemini 1.5 Pro\_text  & 100.0 & 98.0 & 100.0 & 100.0 & 98.0 & 100.0 & 96.0 & 94.0 & 100.0 & 98.0 \\
                              & Claude 3.5 Sonnet\_text     & 100.0 & 96.0 & 96.0 & 98.0 & 100.0 & 100.0 & 98.0 & 94.0 & 96.0 & 100.0 \\
\bottomrule
\end{tabular}
}
\end{adjustbox}
\caption{Value preference outcomes across different models and input settings on Value-Spectrum. The settings include direct multi-modal responses from VLMs and combinations of image descriptions generated by different VLMs with LLMs.}
\label{table:preferences_by_setting}
\end{table}

\end{document}